\pdfoutput=1

\documentclass[11pt]{article}

\usepackage[]{acl}

\usepackage{times}
\usepackage{latexsym}

\usepackage[T1]{fontenc}

\usepackage[utf8]{inputenc}

\usepackage{microtype}

\usepackage{inconsolata}

\usepackage{caption,subcaption,graphicx}
\usepackage{comment}
\usepackage{multirow, multicol}
\usepackage{amsmath,amsfonts,amssymb,bm}
\usepackage{booktabs}
\usepackage{xspace}
\usepackage{enumitem}
\usepackage{setspace}
\usepackage{color, colortbl}
\usepackage{makecell}

\newcommand{\model}{\textsc{D-Pruner}}

\definecolor{amber}{rgb}{1.0, 0.75, 0.0}

\title{Pruning as a Domain-specific LLM Extractor}

\author{\stepcounter{footnote}Nan Zhang$^{\clubsuit}$\thanks{Work done as a Research Intern at NEC Labs America.} \quad Yanchi Liu$^\diamondsuit$ \quad Xujiang Zhao$^\diamondsuit$ \quad Wei Cheng$^\diamondsuit$ \\ {\bf Runxue Bao$^\diamondsuit$ \quad Rui Zhang$^{\clubsuit}$ \quad Prasenjit Mitra$^{\clubsuit}$ \quad Haifeng Chen$^\diamondsuit$} \\
  $^{\clubsuit}$The Pennsylvania State University \quad $^\diamondsuit$NEC Labs America \\
  \texttt{\{njz5124,rmz5227,pmitra\}@psu.edu}\\ \texttt{\{yanchi,xuzhao,weicheng,rbao,haifeng\}@nec-labs.com}}

\begin{document}
\maketitle
\begin{abstract}
Large Language Models (LLMs) have exhibited remarkable proficiency across a wide array of NLP tasks. However, the escalation in model size also engenders substantial deployment costs. While few efforts have explored model pruning techniques to reduce the size of LLMs, they mainly center on general or task-specific weights. 
This leads to suboptimal performance due to lacking \emph{specificity} on the target domain or \emph{generality} on different tasks when applied to domain-specific challenges.
This work introduces an innovative unstructured dual-pruning methodology, \model{}, for domain-specific compression on LLM. It extracts a compressed, domain-specific, and task-agnostic LLM by identifying LLM weights that are pivotal for general capabilities, like linguistic capability and multi-task solving, and domain-specific knowledge.
More specifically, we first assess general weight importance by quantifying the error incurred upon their removal with the help of an open-domain calibration dataset. 
Then, we utilize this general weight importance 
to refine the training loss, so that it preserves generality when fitting into a specific domain. Moreover, by efficiently approximating weight importance with the refined training loss on a domain-specific calibration dataset, we obtain a pruned model emphasizing \emph{generality} and \emph{specificity}. Our comprehensive experiments across various tasks in healthcare and legal domains show the effectiveness of \model{} in domain-specific compression. Our code is available at \url{https://github.com/psunlpgroup/D-Pruner}.
\end{abstract}

\section{Introduction}
Large Language Models (LLMs) such as the GPT family~\citep{brown2020language} and the LLaMA family \citep{touvron2023llama} have exhibited remarkable advancements across a diverse spectrum of NLP tasks. 
However, the substantial size of LLMs engenders cost-intensive deployment in real-world applications and renders them unsuitable for scenarios necessitating efficient inference and low latency \citep{bai2024beyond}. 
Recently, model pruning techniques have been successfully applied to language models \citep{han2015learning, xia2022structured, frantar2023sparsegpt}. These methods aim to yield a compact language model characterized by a significantly reduced parameter count, which is cost-efficient for deployment. However, most of them target relatively small language models, and only a few focus on LLMs \citep{frantar2023sparsegpt, ma2023llm, sun2023simple, xia2023sheared}.
Moreover, the existing strategies mainly center on general or task-specific weights, leading to suboptimal performance due to lacking \emph{specificity} on the target domain or \emph{generality} on different tasks when applied to domain-specific challenges.
Here \textbf{\emph{generality}} refers to the general capabilities of an LLM such as language understanding and generation, and multi-task solving, and \textbf{\emph{specificity}} refers to the capability of an LLM to understand domain-specific knowledge.

\begin{figure}[t!]
    \centering
    \includegraphics[width=0.5\textwidth]{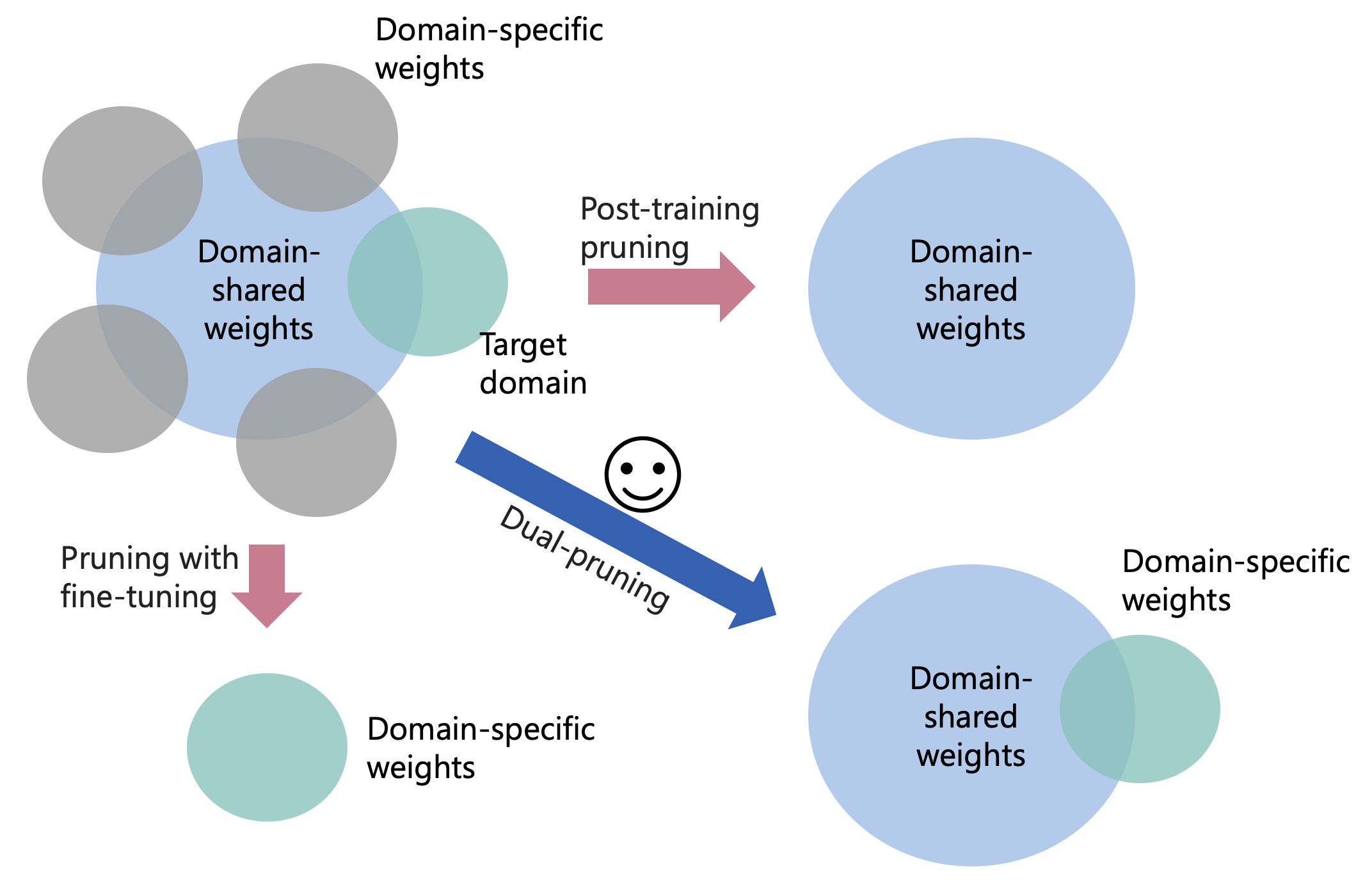}
    \caption{Different types of pruning methods. An LLM is composed of domain-shared weights and domain-specific weights. Post-training pruning focuses on domain-shared weights for generality, pruning with fine-tuning focuses on domain-specific weights for specificity, and our dual-pruning method preserves weights pivotal for both generality and specificity.}
    \label{fig:prune_types}
    \vspace{-0.3cm}
\end{figure}

As shown in \autoref{fig:prune_types}, the weights in an LLM work together to support its general capabilities and to store various domain knowledge. The domain-shared weights (or general weights) empower the LLM with linguistic and multi-task solving prowess akin to human language usage and thinking. The domain-specific weights (or domain weights) are pivotal for endowing the LLM with domain-specific expertise mirroring that of domain experts.
However, the current pruning methods mainly focus on preserving general or task-specific weights, which may not be enough to deal with domain-specific problems.
For example, post-training pruning methods \citep{frantar2023sparsegpt} assume the model is optimized and prune unimportant weights based on an open-domain calibration dataset. This leads to a pruned model that focuses on model generality with domain-specific weights not considered. On the other hand, pruning with fine-tuning methods \citep{ma2023llm} utilizes gradients during fine-tuning on a specific task to estimate the importance of parameters. 
As a result, the pruned model focuses on the model specificity while decreasing the linguistic and multi-task solving capabilities, compromising the LLM's capacity as a versatile task-agnostic solver.

To this end, this study introduces a novel dual-pruning approach, \model{}, for domain-specific unstructured pruning on LLMs, which aims to extract a domain-specific LLM from the foundation LLM. This extracted model is able to solve different tasks in the target domain and facilitates further domain-specific fine-tuning. 
\model{} is designed to harness calibration data for guiding LLM pruning processes while preserving generality and specificity for multi-task solving and domain challenges. The resulting compressed LLM can be seamlessly adapted to the target domain, enabling deployment with limited computing resources. 
Specifically, \model{} adeptly captures and retains both general and domain parameters while selectively eliminating insignificant model parameters. This mechanism comprises the following steps: firstly, a general weight importance module operates to assess the significance of model parameters for general capabilities. Subsequently, we propose an updated training loss function based on the autoregressive training objective for the next token prediction by integrating the general importance as a regularization term. This way, we identify weights contributing to both generality and domain specificity when training on a domain calibration dataset. Then, with the updated loss function, we compute the weight importance leveraging gradients without updating the model. Moreover, an approximation algorithm, empirical Fisher \citep{martens2020new, sung2021training}, is utilized to compute the weight importance efficiently for pruning. 


We evaluate the performance of \model{} on LLaMA2~\citep{touvron2023llama}, a widely adopted open-source LLM. 
Our experimental findings demonstrate that \model{} exhibits remarkable efficiency in the extraction of sparse domain networks from pre-trained LLMs, with a limited amount of calibration data provided. Remarkably, \model{} achieves comparable results to the full dense model while achieving 50\% sparsity, surpassing the performance of alternative pruning techniques across diverse domain-specific datasets in healthcare and legal domains encompassing language comprehension, question answering, and summarization tasks.

\section{Related Work}

Model compression involves transforming a large, resource-intensive model into a compact version suitable for low-resource deployment \cite{deng2020model, zhu2023survey}. There are mainly three techniques for model compression, which are pruning, knowledge distillation, and quantization.

\paragraph{Pruning.}
Pruning techniques in neural networks can be broadly classified into structured pruning and unstructured pruning \citep{xia2022structured, sanh2020movement, du2021robustness}. \textit{Structured pruning} entails the removal of entire network components, such as channels or layers, guided by specific criteria, while maintaining the overall network architecture. In contrast, \textit{unstructured pruning} targets individual weights, leading to an irregular sparse structure. 

While numerous attempts have been made to prune language models of relatively small scales, such as BERT \citep{kenton2019bert}, scant attention has been devoted to pruning LLMs containing billions of parameters. These larger models possess 100-1000 times more weights, rendering the pruning task significantly more challenging. SparseGPT \citep{frantar2023sparsegpt}, a post-training method for Large Language Models (LLMs), lacks the capability to identify crucial weights tailored to specific domains or tasks as it refrains from fine-tuning. On the other hand, LLM-Pruner \citep{ma2023llm} employs gradient-based techniques for pruning. However, it falls short in identifying pivotal weights essential for domain-shared knowledge, resulting in pruned models that lack the desired level of generality.

The existing pruning methods either focus on general or domain-specific weights, yet none of them consider preserving both at the same time. To the best of our knowledge, we are the first to work on pruning LLMs while preserving weights important to both generality and specificity.

\paragraph{Knowledge Distillation.}
Knowledge Distillation (KD) has emerged as a powerful technique, drawing considerable interest for its ability to augment model performance and enhance generalization capacities \citep{hinton2015distilling, zhu2023survey}. At its core, KD revolves around the transfer of expertise from a complex model, referred to as the ``teacher model'', to a simplified counterpart known as the ``student model''. This intricate process of knowledge transfer aims to distill the profound insights encapsulated within the teacher models, condensing them into a more concise and efficient representation within the student models.

While KD has been proven a powerful tool for model compression, it needs specific downstream tasks and a large amount of data for the student models to learn from the teacher models. Thus, the output that student models produce mainly focuses on a specific task and loses the generality capability. KD generally sets higher requirements on data availability and computation budgets (e.g., GPU memory) than pruning.

\paragraph{Quantization.}
In the realm of model compression, quantization has emerged as a widely embraced technique to alleviate the storage and computational challenges inherent in deep learning models \citep{guo2020accelerating,dettmers20218,dettmers2022llm,dettmers2023qlora}. Conventional model representations rely on floating-point numbers, but quantization converts them into integers or discrete forms. This transformation leads to substantial reductions in storage requirements and computational complexities. While a certain degree of precision loss is inevitable, carefully designed quantization methods can achieve significant model compression with minimal accuracy degradation.
Although challenges remain, such as maintaining model interpretability and addressing task-specific intricacies, the current body of research establishes a robust groundwork for ongoing advancements in LLM quantization, which could be complementary to LLM pruning.

\section{Methodology}
\label{gen_inst}
To preserve both generality and specificity on the pruned model, our dual-pruning method \model{} considers weights important to both generality and specificity during training on a calibration dataset. Note we only use the weight gradient generated from the training process but do not update the model weights. Our model is pruned in a task-agnostic fashion (e.g., we adopted a pre-training objective, next token prediction, as a part of training loss) so that the pruned model can solve different tasks in the target domain. 

\model{} comprises the following steps: firstly, a general weight locating module operates to assess the significance of model parameters for general understanding (Section~\ref{sec:general_weight}). Subsequently, an updated loss function for the training process is proposed by integrating the general weight importance as a regularization term. This way, we identify weights contributing to both general and domain knowledge (Section~\ref{sec:updated_loss}). Finally, with the updated loss function, we compute the weight gradients on a small domain calibration dataset without updating the model and approximate our dual-pruning weight importance by utilizing the empirical Fisher index \citep{sung2021training} for pruning (Section~\ref{sec:dual_pruning}).

Our method concentrates on unstructured pruning in a layer-by-layer manner for the Transformers model. We consider query, key, value, and output projections of all self-attention layers and gate \citep{liu2021pay}, down, and up projections of all MLP (multilayer perceptron) layers for pruning.

\subsection{General Weight Importance}
\label{sec:general_weight}
The first step of our method involves locating important weights in terms of general knowledge. Following the same hypothesis as \citet{frantar2023sparsegpt}, we assume that an important weight will cause a larger increase in loss value than those less important ones if it is pruned (set to $0$) during training. Formally, if a dataset of the open-domain calibration $\mathcal{D}_g = \{x_j, y_j\}^{N}_{j=1}$ with size $N$ 
is used for training and $W$ stands for weight matrices of a model, the importance of each weight at index $m$, denoted as $I_{W^m}$, can be approximated using Taylor series as shown
by \citet{lecun1989optimal}:
\begin{equation}
\begin{aligned}    
\label{eq1}
    &I_{W^m} = |\mathcal{L}(\mathcal{D}_g) - \mathcal{L}_{W^m=0}(\mathcal{D}_g)| \\
    &= |\frac{\partial \mathcal{L}(\mathcal{D}_g)}{\partial W^m}W^m + \frac{1}{2}W^m H_{mm} W^m \\
    &+ O(||W^m||^3)|
\end{aligned}
\end{equation}
where $H$ denotes the Hessian matrix, and $\mathcal{L}$ is the cross-entropy loss. For a model that is sufficiently trained to a local minimum on its loss curvature (e.g., pretrained foundational language models such as LLaMA), the classic Optimal Brain Surgeon \citep{hassibi1993optimal} further approximates the importance of $W^m$ as:
\begin{equation}
\label{eq2}
    \varepsilon^m = \frac{1}{2}\frac{(W^m)^2}{[H^{-1}]_{mm}}
\end{equation} 
$\varepsilon^m$ can also be viewed as the error caused by removing the weight $W^m$. We compute $\varepsilon^m$ for all the weights subject to pruning and construct a matrix of importance scores $G$ with respect to general domains that have the same dimension as $W$.

\subsection{Updated Loss with Regularization}
\label{sec:updated_loss}

To identify the weights that are important in both general and domain-specific knowledge, we modify the original loss function of LLM training. In LLM training, cross-entropy loss is used in the next token prediction task \citep{radford2018improving}. Similar to \citet{thompson2019overcoming}, we add a regularization term to constrain the change of important general weights found in the first step. Suppose that there are $M$ number of prunable weights in total. To train on a domain-specific calibration dataset $\mathcal{D}_s = \{x_j, y_j\}^{P}_{j=1}$, we add the proposed regularization term on top of the next token prediction loss $\mathcal{L}_{next}$ to obtain our final training objective:
\begin{equation}
    \mathcal{L}_{\text{ours}}  = \mathcal{L}_{\text{next}} + \lambda \sum_{m=1}^{M} G^m ({W^m}'-W^m)^2
\end{equation} 
where $G^m$ is the general weight importance, ${W^m}'$ denotes the updated weight value of $W^m$, $\lambda$ is a hyperparameter, and the second term on the right is $\mathcal{L}_{\text{regular}}$.

In practice, the direct calculation of this regularization term in the forward pass is computationally expensive for two reasons: (1) it involves both $W^m$ and $G^m$ which are very large, and (2) gathering updated model parameters (${W^m}'$) in a partitioned \citep{rasley2020deepspeed} or sharded \citep{zhao2023pytorch} system is inefficient. Based on the recent success of applying gradient descent on full fine-tuning of LLMs \citep{lv2023full}, we choose to use gradient descent to optimize parameters. Therefore, at a learning rate $\alpha$, denoting the gradient of each parameter with respect to $\mathcal{L}_{\text{next}}$ as $g_{\text{next}}^m$, we reduce the regularization term to:
\begin{equation}
\begin{aligned}
    \mathcal{L}_{\text{regular}} &= \sum_{m=1}^{M} G^m ({W^m}'-W^m)^2 \\
    &= \lambda \sum_{m=1}^{M} G^m (W^m - \alpha g_{\text{next}}^m - W^m)^2 \\
    &= \lambda \sum_{m=1}^{M} \alpha^2 G^m (g_{\text{next}}^m)^2
\end{aligned}
\end{equation}


During the backward pass, optimizing this regularization term requires second-order derivatives, which indicates that
Hessian matrices ($H$) are needed. Directly computing the Hessian matrices is infeasible for such a large number of parameters. Therefore, we use the Fisher information matrix to approximate the diagonal of the Hessian \citep{sung2021training}. And the Fisher information matrix can be further approximated by the 
average of the squared gradient of the model’s prediction over $P$. We write the gradient of the regularization with respect to every parameter matrix in a finer granularity:
\begin{equation}\label{mathrefs}
    \frac{\partial \mathcal{L}_{\text{regular}}}{\partial W^m} \approx 2\lambda\alpha^2 G^m g_{\text{next}}^m H_{mm}
\end{equation}


\begin{equation}
H_{mm} \approx \frac{1}{P}\sum_{j=1}^{P} (g_{\text{next}}^m(x_j, y_j))^2
\end{equation}

We directly compute $\frac{\partial \mathcal{L}_{\text{regular}}}{\partial W}$ via Equation~\ref{mathrefs} above
instead of relying on PyTorch backward pass to maximize computing efficiency. The final gradient computation of our regularized loss function is shown below:

\begin{equation}
    \frac{\partial \mathcal{L_{\text{ours}}}}{\partial W^m} = \frac{\partial \mathcal{L_{\text{next}}}}{\partial W^m} + \frac{\partial \mathcal{L_{\text{regular}}}}{\partial W^m}
\end{equation}




\begin{table*}[]
\resizebox{\textwidth}{!}{%
\begin{tabular}{@{}lccccccc@{}}
\toprule
 & InternalMed\_Harrison & MedNLI & PubMedQA & HQS & MultiLegalPile & CaseHOLD & BillSum \\ \midrule
Domain & Healthcare & Healthcare & Healthcare & Healthcare & Legal & Legal & Legal \\
Task / Type & Generation & NLI & QA & Summarization & Generation & QA & Summarization \\
\# Instances in Test & 300 & 1422 & 500 & 100 & 300 & 200 & 200 \\ 
Metrics & Perplexity & Accuracy & Macro-F1 & ROUGE & Perplexity & Macro-F1 & ROUGE \\ \bottomrule
\end{tabular}%
}
\caption{Details of each dataset that we use for model evaluation.}
\label{tab:dataset}
\end{table*}

\subsection{Dual-pruning Importance Score}
\label{sec:dual_pruning}

Finally, we calculate the dual-pruning importance score of each weight, and unimportant weights can be pruned according to their importance. 
We use Equation~\ref{eq1} for importance estimation instead of Equation~\ref{eq2}, because our model has not converged to an optimum on the target domain. However, direct computation of the Hessian matrix in 
Equation~\ref{eq2} is infeasible since it involves $O(M^2)$ complexity for each weight update. Therefore, we also 
leverage \citet{sung2021training} to approximate the diagonal of the Hessian, and the final importance score $S^m$ can be defined as:
\begin{equation}
\begin{aligned}
    S^m &\approx |\frac{\partial \mathcal{L_{\text{ours}}}(\mathcal{D}_s)}{\partial W^m}W^m + \frac{1}{2} [\frac{\partial \mathcal{L_{\text{ours}}}(\mathcal{D}_s)}{\partial W^m} W^m]^2 \\
    &+ O(||W^m||^3)|
\end{aligned}
\end{equation}
Here $O(||W^m||^3)$ can be neglected according to the quadratic approximation \cite{lecun1989optimal}.
Note the calculation of $S^m$ considers both general and domain-specific knowledge via our regularized training objective. 
Combining both regularization and importance estimation via empirical Fisher approximation, our method expects to conduct pruning that maintains weights important to both general and domain-specific knowledge, thus preserving generality and specificity.
And these importance scores are used to guide our pruning decisions. For example, if we set the sparsity level to be 50\%, weights that have the smallest 50\% of importance scores in each layer will be pruned.

\section{Experiment Setup}
\label{sec:exp}
We evaluate \model{} on two knowledge-intensive domains, which are healthcare and legal. For model generality under domain-specific challenges, we evaluate the linguistic capability using domain text generation, and evaluate the multi-task solving capability on different domain tasks, i.e., natural language inference (NLI), question answering (QA), and summarization. Since we use domain datasets, the model specificity on domains can also be evaluated. In addition, we fine-tune the pruned model on domain datasets to further evaluate the generality and specificity.

We evaluate \model{} on the LLaMA2 model family, which is the most used open-source LLM. We mainly apply our pruning method and baseline methods to LLaMA2-7B and LLaMA2-13B to show our results. Our method can also be easily applied to other LLMs with different sizes and architectures. For instance, Appendix~\ref{sec:exp_bloom} shows further experiment on BLOOM model \citep{le2022bloom}.

\subsection{Iterative blocking}
\label{subsec:iterative_b}
Motivated by \citet{frantar2023sparsegpt}, we perform experiments (in Table~\ref{tab:7&13}) on \model{} with and without iterative blocking. Iterative blocking means to make pruning decisions for every fixed number ($B_s$) of columns within a weight matrix. In other words, instead of selecting a single pruning mask for an entire weight matrix, a pruning sub-mask is selected for every $B_s$ columns to reach overall sparsity level. We set $B_s = 128$ for weight matrices with the smallest number of columns and increase $B_s$ for those with more columns. Except Table~\ref{tab:7&13}, \model{} in other tables does not adopt iterative blocking.

\subsection{Datasets and Evaluations}
\textbf{Datasets.}
Table~\ref{tab:dataset} shows the details of each dataset that we used.
Specifically, for healthcare, we select a medical textbook InternalMed\_Harrison \citep{bigby1988harrison}, MedNLI \citep{romanov-shivade-2018-lessons}, PubMedQA \citep{jin-etal-2019-pubmedqa}, and Health Question Summarization (HQS) from the MEDIQA 2021 shared task 1 \citep{ben-abacha-etal-2021-overview,ben-abacha-demner-fushman-2019-summarization} as domain datasets. For legal domain, we select MultiLegalPile \citep{niklaus2023multilegalpile}, CaseHOLD \citep{zhengguha2021}, and BillSum \citep{kornilova-eidelman-2019-billsum}. As for open-domain calibration data, we extract text from C4 dataset \citep{2019t5}.

To construct our domain-specific calibration data, we select training instances from MedNLI, PubMedQA, and HQS at a ratio of 20\%/60\%/20\% and from CaseHOLD and BillSum at a ratio of 50\%/50\%. These ratios are determined based on the difficulties and training sizes of these benchmarks. Both NLI and QA tasks that we adopt are asking models to perform classification. We experiment with different sizes of the domain-specific calibration dataset and find a size of 1000 achieves the best trade-off in terms of pruning efficiency and effectiveness for both domains. For model evaluation, besides using the test instances of those benchmarks, we leverage InternalMed\_Harrison and MultiLegalPile for perplexity evaluation. 300 paragraphs are selected from each data source to form the test set of perplexity. Note that we use a subset of all the test examples of CaseHOLD and BillSum, since these two benchmarks are significantly larger in size and their individual instance tends to be longer.


\noindent\textbf{Evaluation Metrics.} We first evaluate the linguistic capability of pruned models on InternalMed\_Harrison and MultiLegalPile using perplexity. We then evaluate the multi-task solving capability and domain specificity on different domain tasks. Specifically, we choose accuracy metric for NLI task (MedNLI), macro-F1 for QA tasks (PubMedQA and CaseHOLD), and ROUGE scores \citep{lin-2004-rouge} for summarization tasks (HQS and BillSum). 




\begin{table*}[t!]
\resizebox{\textwidth}{!}{%
\begin{tabular}{@{}lccccccccccc@{}}
\toprule
\multirow{2}{*}{Model} & \multicolumn{6}{c}{Healthcare} & \multicolumn{5}{c}{Legal} \\
\cmidrule(lr){2-7} \cmidrule(lr){8-12}
 & Perplexity & MedNLI & PubMedQA & R1 & R2 & RL & Perplexity & CaseHOLD & R1 & R2 & RL \\ \midrule
 & \multicolumn{11}{c}{LLaMA2-7B} \\
 \midrule
 Dense & 5.49 & 37.62 & 23.77 & 22.51 & 7.18 & 19.50 & 2.26 & 28.82 & 32.64 & 18.32 & 26.48 \\
 Magnitude \citep{han2015learning} & 16.08 & 33.90 & 28.29 & 9.60 & 1.63 & 8.09 & 8.64 & 23.84 & 7.84 & 2.21 & 6.13\\ 
 LLM-Pruner \citep{ma2023llm} & 88.25 & 33.90 & 22.34 & 5.52 & 0.30 & 5.45 & 32.22 & 13.59 & 6.76 & 0.72 & 5.40 \\
 SparseGPT \citep{frantar2023sparsegpt} & \textbf{6.39} & 33.47 & 36.22 & 22.60 & 7.68 & 19.13 & \textbf{2.62} & 28.41 & 32.68 & \textbf{18.89} & 26.19  \\
 \model{} (w/ iterative blocking) & 7.07 & 34.53 & \textbf{45.38} & 24.72 & 8.87 & 21.09 & 2.70 & \textbf{30.56} & \textbf{33.77} & 18.53 & \textbf{26.25}\\
 \model{} (w/o iterative blocking) & 6.96 & \textbf{34.81} & 42.40 & \textbf{25.05} & \textbf{9.65} & \textbf{22.34} & 2.72 & 26.14 & 32.14 & 18.42 & 26.14 \\
 \midrule & \multicolumn{11}{c}{LLaMA2-13B} \\
 \midrule
 Dense & 5.20 & 35.02 & 40.54 & 19.26 & 5.80 & 16.40 & 2.12 & 28.89 & 35.34 & 21.19 & 27.82 \\
 Magnitude \citep{han2015learning} & 6.59 & \textbf{36.71} & 45.12 & 19.60 & 5.01 & 16.33 & 2.81 & 21.95 & 29.90 & 16.94 & 24.51 \\
LLM-Pruner \citep{ma2023llm} & 23.95 & 34.39 & 17.37& 7.60& 1.24 & 7.00 & 12.16 & 13.46 & 17.21 & 3.08 & 12.37 \\
SparseGPT \citep{frantar2023sparsegpt} & \textbf{5.77} & 34.39 & 52.65 & 22.25 & \textbf{8.35} & 19.19 & \textbf{2.39} & \textbf{28.62} & 33.68 & 19.35 & 27.60 \\
\model{} (w/ iterative blocking) & 6.30 & 34.88 & \textbf{52.86} & 20.56 & 6.95 & 17.85 & 2.40 & 28.30 & 33.83 & 20.51 & 27.56 \\
\model{} (w/o iterative blocking) & 6.16 & 35.16 & 50.87 & \textbf{23.99} & 7.78 & \textbf{20.04} & 2.40 & 27.27 & \textbf{35.77} & \textbf{21.81} & \textbf{28.42} \\
 \bottomrule
\end{tabular}%
}
\caption{Overall results when candidate models (at 50\% sparsity) are tested on two domains. The best scores are in \textbf{bold} except the ones from the dense models. Note that the ROUGE scores reported in the healthcare domain correspond to HQS dataset while those in the legal domain correspond to BillSum. Perplexity in healthcare is tested on InternalMed\_Harrison and perplexity in legal is tested on MultiLegalPile.}
\label{tab:7&13}
\end{table*}

\subsection{Baselines}
We compare our method with a variety of LLM pruning baselines. All methods are applied to the same foundation model (either 7B of 13B of LLaMA2) for fair comparisons. As an ablation study, we also evaluate an unstructured pruning method using weight gradient by removing the regularization term in the training loss of \model{}.

\begin{itemize}[noitemsep,topsep=0pt,parsep=0pt,partopsep=0pt,leftmargin=*]
\item \textbf{Magnitude pruning} prunes weights based on their magnitudes \citep{han2015learning}. We follow the standard practice of magnitude pruning on language models, where weights are compared layer-wise. Magnitude pruning is a simple and robust baseline that has been demonstrated to outperform many other pruning methods.

\item \textbf{LLM-Pruner} is a structured pruning method using weight gradient to evaluate weight importance \citep{ma2023llm}. A calibration dataset is used for its gradient calculation, so we combine both open-domain (C4) and domain-specific calibration data when we use LLM-Pruner.

\item \textbf{SparseGPT} is an unstructured post-training pruning method \citep{frantar2023sparsegpt}. It uses an efficient weight update procedure that iterates between weight removal and weight update at each layer. It also uses a calibration dataset for approximation. Thus, similarly to \model{} and LLM-Pruner, we use open-domain and domain-specific calibration data for fair comparisons.
\end{itemize}

Moreover, for all the baseline methods, we continue to fine-tune their pruned models using LoRA \citep{hu2021lora} on all the datasets together (NLI, QA, and summarization data combined) in each domain and then test the fine-tuned model on the datasets in Table~\ref{tab:dataset}. We only use the default open-domain calibration dataset for the pruned models of LLM-Pruner and SparseGPT at this step, because these models will eventually undergo LoRA fine-tuning. Data instances of our fine-tuning dataset follow the Alpaca \citep{alpaca} template so that models are trained to predict the responses. Specifically, for healthcare, we have 7000, 7000, and 1000 training instances from MedNLI, PubMedQA, and HQS, respectively. For legal domain, we have 13000 training instances from CaseHOLD and 2000 from BillSum.


\subsection{Implementation Details}
We perform prompt engineering in a zero-shot setting before prompting a series of models. The finalized prompt is kept the same across all candidate models on one task to ensure fairness. The hyperparameters used by different models are in Appendix~\ref{sec:hyper}.

\section{Results and Analysis}
\label{sec:results}
Our results and analysis aim to answer the following research questions:
\begin{itemize}[noitemsep,topsep=0pt,leftmargin=0.4cm]
    \item RQ 1: How does \model{} compare against other pruning baselines (\ref{sec:overall_results})?
    \item RQ 2: What are the performance of all candidate models after LoRA fine-tuning (\ref{sec:ft})?
    \item RQ 3: As an important contribution of \model{}, is dual-pruning an effective method of compressing LLM (\ref{sec:overall_results}, \ref{sec:ablation}, and \ref{sec:mask})?
    \item RQ 4: How does \model{} perform under different sparsity levels or different sizes of domain-specific calibration data (\ref{sec:domain_calibration})?
\end{itemize}

\subsection{Overall Results}
\label{sec:overall_results}
Our overall results for the two domains are presented in Table~\ref{tab:7&13}. All models are pruned to 50\% sparsity level except the dense one.

\begin{table*}[t!]
\resizebox{\textwidth}{!}{%
\begin{tabular}{@{}lccccccccccc@{}}
\toprule
\multirow{2}{*}{Model (Fine-tuned with LoRA)} & \multicolumn{6}{c}{Healthcare} & \multicolumn{5}{c}{Legal} \\
\cmidrule(lr){2-7} \cmidrule(lr){8-12}
 & Perplexity & MedNLI & PubMedQA & R1 & R2 & RL & Perplexity & CaseHOLD & R1 & R2 & RL \\ \midrule
 & \multicolumn{11}{c}{LLaMA2-7B} \\
 \midrule
 Dense & 5.68 & 64.84 & 41.37 & 33.26 & 12.60 & 28.92 & 2.26 & 28.82 & 34.64 & 20.47 & 28.33 \\
 Magnitude \citep{han2015learning} & 8.39 & 62.59 & 23.71 & 32.02 & 12.25 & 29.27 & 7.28 & 25.89 & 17.64 & 8.19 & 14.52\\ 
 LLM-Pruner \citep{ma2023llm} & 44.56 & 58.72 & 26.78 & 22.21 & 6.12 & 20.57 & 215.13 & 14.37 & 7.97 & 0.78 & 6.68 \\
 SparseGPT \citep{frantar2023sparsegpt} & \textbf{6.44} & \textbf{68.85} & 27.37 & 28.97 & 11.27 & 25.93 & 2.86 & 27.31 & 27.79 & 17.55 & 23.74 \\
 \model{} & 6.74 & 61.88 & \textbf{32.58} & \textbf{36.49} & \textbf{13.71} & \textbf{31.85} & \textbf{2.73} & \textbf{27.58} & \textbf{31.00} & \textbf{19.03} & \textbf{25.96} \\
 \bottomrule
\end{tabular}%
}
\caption{Results of fine-tuned candidates models at 50\% sparsity. LoRA fine-tuning is conducted on \model{} without iterative blocking.}
\label{tab:lora}
\end{table*}

\paragraph{Improvement on NLI and QA}
\model{} delivers consistent score improvement on NLI and QA tasks when it is compared against baselines based on LLaMA2-7B and LLaMA2-13B. With two exceptions, variants of \model{} based on the inclusion and exclusion of iterative blocking outperform baselines on 4 out of 6 cases when classification is performed (MedNLI, PubMedQA, and CaseHOLD on both 7B and 13B LLaMA2) in Table~\ref{tab:7&13}. It is clear to see that magnitude pruning and SparseGPT are generally stronger models than LLM-Pruner. The dense model sometimes has worse scores than others across 7B and 13B LLaMA2, which indicates that scaling parameters of a pre-trained language model does not necessarily increase the performance on a single benchmark on NLI and QA. We can see that iterative blocking generally yields better scores on these classification tasks such as reaching 30.56 F1 score on CaseHOLD based on LLaMA2-7B, which is a significant improvement over baselines and \model{} without it. Thus, we recommend to adopt iterative blocking on the classification tasks when strong domain knowledge is required.

\paragraph{Improvement on Summarization}
\model{} presents the strongest summarization performance. The most exciting thing is that its ROUGE scores are mostly higher than the dense ones. We notice the top summarization performance of LLaMA2-13B-based models on HQS is lower than that of LLaMA2-7B-based models, which is counterintuitive. According to the state-of-the-art of HQS \cite{zhang-etal-2023-famesumm,he-etal-2021-damo}, we find that \model{} is close to the best ROUGE scores produced by single systems, so we consider that this dataset is relatively simple. Thus, our LLaMA2-7B-based models seem to find an upper limit of ROUGE given the existing reference summaries, so going from 7B to 13B incurs a small performance degradation on dense model, SparseGPT, and \model{}. The strong summarization performance of \model{} on both domains demonstrates its usability as an efficient and domain-specific language model. As for iterative blocking, \model{} without it generally has better perplexity and summarization performance. However, considering the exception in the legal domain based on LLaMA2-7B, we recommend to check perplexity scores on the validation data when deciding whether to use iterative blocking for perplexity and summarization assessment.

\paragraph{Improvement on Perplexity}
\model{} has the second best perplexity scores on healthcare and legal domains across 7B and 13B LLaMA2. These scores reflect the strong linguistic capabilities of SparseGPT and \model{} when they encounter knowledge-intensive domains. \model{} does not surpass SparseGPT on perplexity metric, and the reason might come from the fine-tuning pipeline \citep{lv2023full} we use. \citet{lv2023full} is a full-parameter fine-tuning pipeline that aims towards GPU memory efficiency, so its effectiveness on a specific metric might be compromised. Moreover, we suspect that the data we use from InternalMed\_Harrison and MultiLegalPile may be closer to the general domain both semantically and syntactically. Since SparseGPT prunes LLM mainly based on generality, it has better perplexity scores than ours.

\subsection{Performance After Fine-tuning}
\label{sec:ft}
Table~\ref{tab:lora} shows the results of fine-tuned candidate models at 50\% sparsity. Similar to the performance discussed above, \model{} always delivers the best summarization scores and mostly presents the best classification results after fine-tuning, which demonstrates that fine-tuning can further improve the pruning performance of our method. For most models, macro-F1 on PubMedQA decreases after fine-tuning, because this test set is imbalanced and models mostly learn to predict the majority class labels. In fact, the accuracies of most models on PubMedQA increase after fine-tuning as shown in Appendix~\ref{sec:acc_pubmed}, so this fine-tuning method still makes a difference. We also do not see too much score improvement for many models on CaseHOLD, since it is a quite challenging task for our experiment setting (e.g., we combine only a small subset of original training data for each task and perform multi-task fine-tuning as discussed in Section~\ref{sec:exp}).

\subsection{Ablation Study}
\label{sec:ablation}
In Table~\ref{tab:regular}, we show that pruning without integrating general domain importance as a regularization term yields suboptimal performance. In other words, this means to remove the consideration of generality. We find perplexities in both domains are higher than pruning with regularization. This demonstrates that our dual pruning mechanism that considers both generality and specificity is able to improve model performance.

\begin{table}[t!]
\resizebox{\columnwidth}{!}{%
\begin{tabular}{lcc}
\toprule
Model & Healthcare perplexity & Legal perplexity \\ \midrule
no regularization & 7.23 & 2.82\\ 
\model{} & \textbf{6.96} & \textbf{2.72} \\ 
\bottomrule
\end{tabular}%
}
\caption{Results of removing the regularization.}
\label{tab:regular}
\end{table}

\begin{table}[t!]
\resizebox{\columnwidth}{!}{%
\begin{tabular}{lcc}
\toprule
Sparsity & Healthcare perplexity & Legal perplexity \\ \midrule
10\% & 5.49 & 2.26\\ 
20\% & 5.52 & 2.27 \\ 
30\% & 5.61 & 2.31 \\ 
40\% & 5.91 & 2.42 \\ 
50\% & 6.96 & 2.72 \\ 
60\% & 15.19 & 4.59 \\ 
70\% & 223.63 & 84.25 \\ 
\bottomrule
\end{tabular}%
}
\caption{Results of changing sparsities on \model{}.}
\label{tab:sparsity}
\end{table}

\begin{table}[t!]
\resizebox{\columnwidth}{!}{%
\begin{tabular}{lcc}
\toprule
\# samples & Healthcare perplexity & Legal perplexity \\ \midrule
100 & 8.18 & 3.34\\ 
500 & 7.15 & 2.97 \\ 
1000 & 6.96 & 2.72 \\ 
1500 & 7.96 & 2.70 \\ 
\bottomrule
\end{tabular}%
}
\caption{Results of trying different sizes of domain-specific calibration data.}
\label{tab:data_size}
\end{table}

\subsection{Effect of Sparsity and Domain Calibration Data}
\label{sec:domain_calibration}
In Table~\ref{tab:sparsity}, it is clear that perplexity keeps increasing when \model{} becomes more sparse, which is expected. Since 50\% sparsity is a good balance between sparsity and performance, we select it to report our performance in Table~\ref{tab:7&13} and \ref{tab:lora}.

Based on Table~\ref{tab:data_size}, we believe setting the size of domain-specific calibration data to 1000 is reasonable. As the last row shows, increasing its size does not always guarantee a performance improvement.

\begin{figure*}[t!]
\centering
\begin{subfigure}[b]{0.4\textwidth}\includegraphics[width=\textwidth]{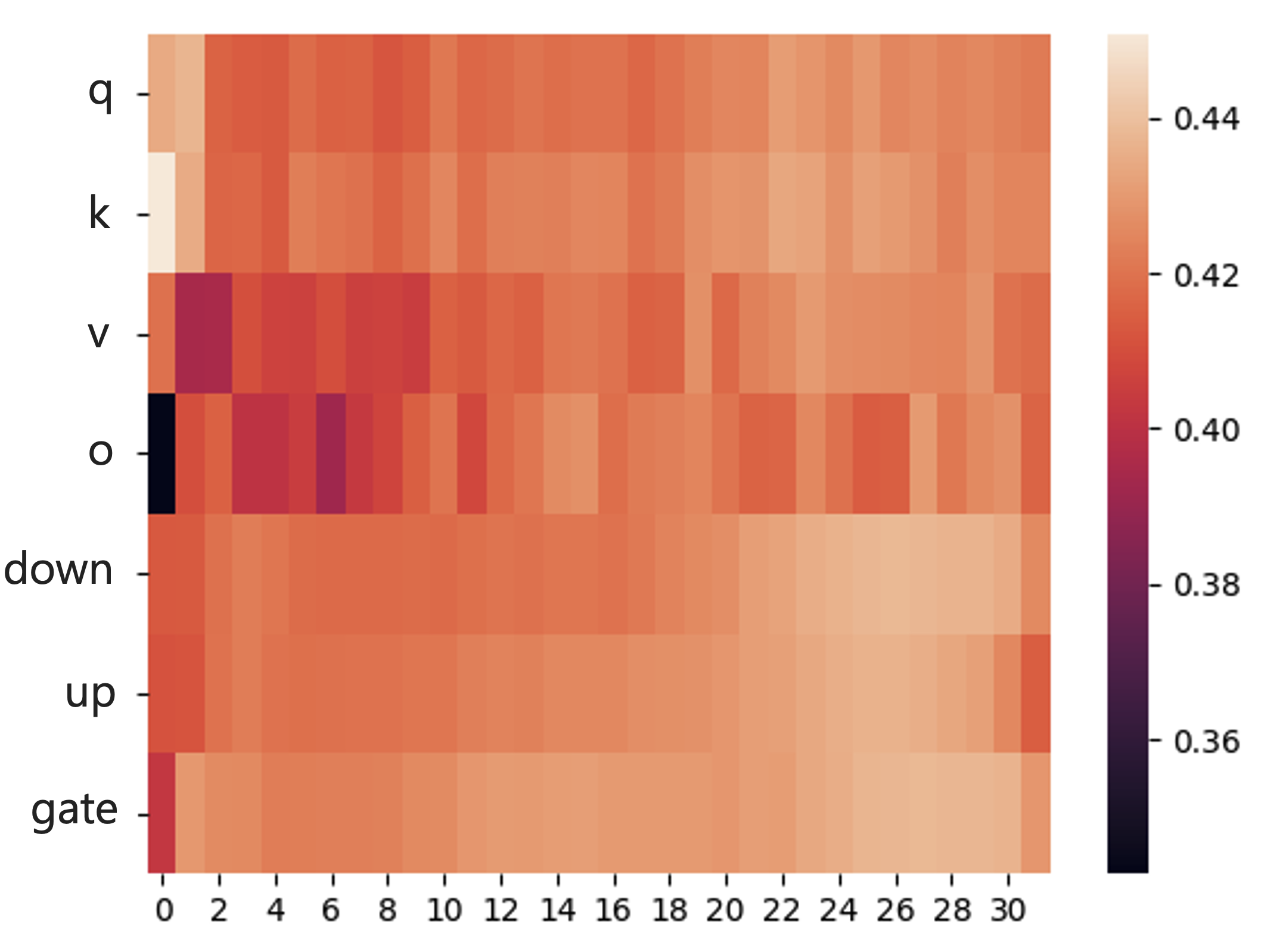}
    \caption{Open-domain vs healthcare domain.} 
    \label{fig:seq2seq_train}
\end{subfigure}
\begin{subfigure}[b]{0.4\textwidth}
\includegraphics[width=\textwidth]{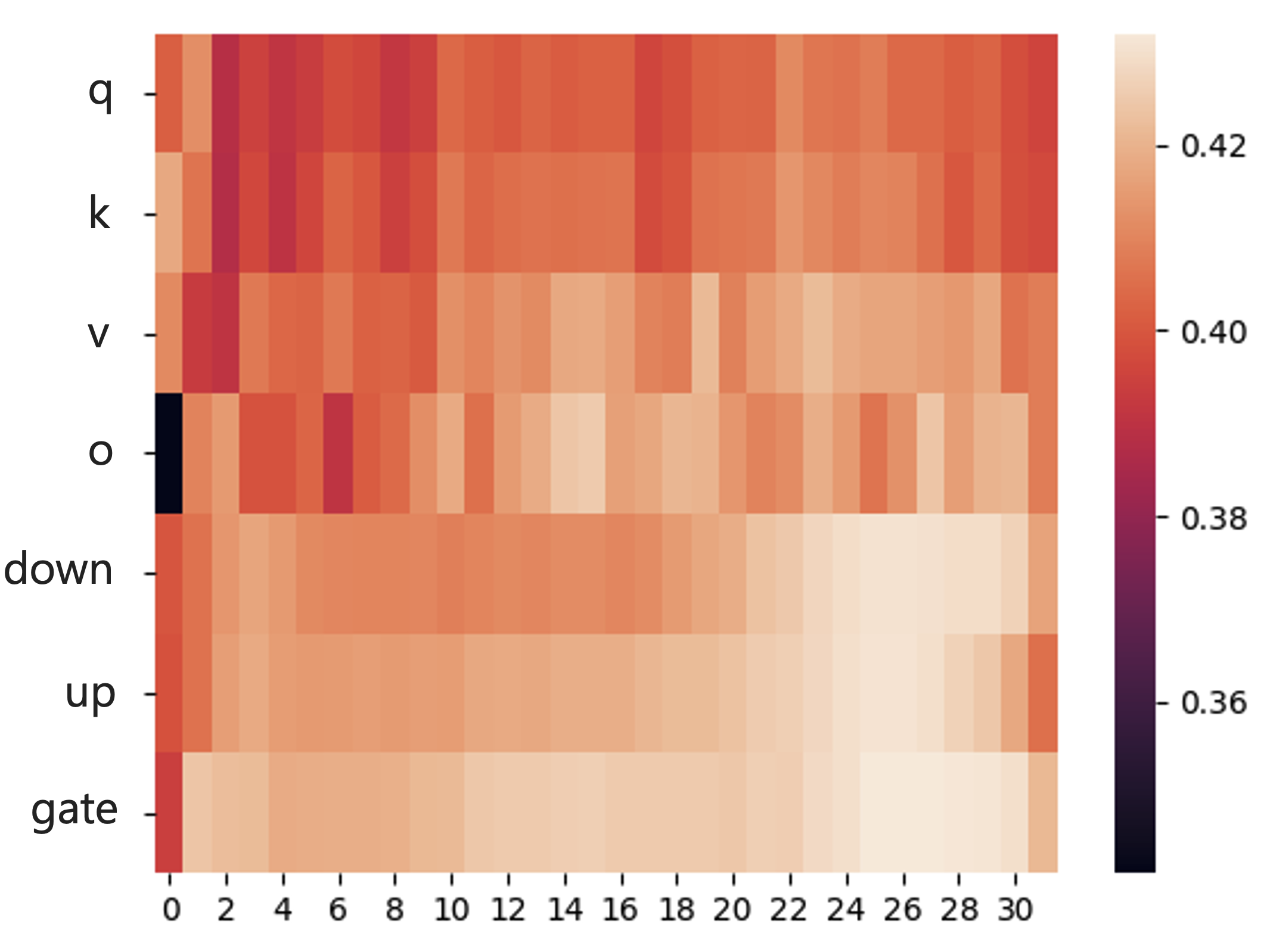}
    \caption{Healthcare domain vs legal domain.}
    \label{fig:seq2seq_test}
\end{subfigure}
\caption{Illustration of mask similarity. It shows that masks for different domains are quite different. The self-attention modules contribute more to specificity, and MLP modules store knowledge that is shared by different domains.}
\label{fig:masksim}
\end{figure*}

\subsection{Mask Similarity}
\label{sec:mask}

To better understand the pruned model on different domains, we compare the similarity of the pruning masks.
In our study on LLaMA2-7B, each generated mask contains 7*32 matrices for 32 layers and 7 projection matrices in the self-attention module (q, k, v, o) and MLP module (down, up, gate) in each layer. For each matrix, we calculate the similarity as the number of shared ``1'' elements (``1'' means weights not pruned) in the two masks divided by the matrix size. Note all the masks are generated in 50\% sparsity.

Figure \ref{fig:masksim} (a) shows the mask similarity between the open-domain and healthcare domain, and \ref{fig:masksim} (b) shows the mask similarity between the healthcare domain and legal domain. 
The results show that the masks are quite different, with shared elements as low as 35\%. 
Generally, the self-attention modules share fewer elements than the MLP modules. This means self-attention modules contribute more to specificity, and MLP modules store knowledge that is shared by different domains.

\section{Conclusion}

We introduce \model{}, an innovative unstructured dual-pruning method for domain-specific compression on LLM. It is able to extract a compressed, domain-specific, and task-agnostic LLM by identifying weights that are pivotal for both generality and specificity.
More specifically, the general weight importance is first assessed by quantifying the error incurred upon their removal with the help of open-domain calibration data. 
Then, we utilize this general weight importance to refine our training loss, so that it considers generality when fitting into a specific domain. Moreover, by efficiently approximating weight importance with the refined training loss on a domain-specific calibration dataset, we obtain a pruned model emphasizing general capabilities and domain-specific knowledge. Our comprehensive experiments across various tasks in different domains show the effectiveness of \model{} in domain-specific pruning.

\section*{Limitations}
Although \model{} presents strong performance in Section~\ref{sec:results}, many of its perplexity scores reach the second place in healthcare and legal domains (dense model is not counted here). Further improving this perplexity is a valuable extension of this paper. 

Another limitation of this work is that \model{} is more memory-intensive than SparseGPT during pruning, since \model{} is based on full-parameter fine-tuning and SparseGPT does not leverage global gradient information. \model{} sets similar memory requirement as LLM-Pruner. As a trade-off, \model{} reaches better performance on most of the metrics. It is also more flexible, since it computes matrices of importance scores without actually sparsifying LLMs. Therefore, researchers can make real-time decisions about the desired sparsity level, and changing the sparsity is very efficient.

\section*{Acknowledgments}
We thank Yusen Zhang, Sarkar Snigdha Sarathi Das, Ranran Haoran Zhang, Xiaoxin Lu, and Ryo Kamoi for the valuable discussions and comments. We also would like to thank the anonymous reviewers for their helpful comments.

\bibliography{anthology,custom}

\appendix

\section{Accuracy Scores on PubMedQA}
\label{sec:acc_pubmed}
In Table~\ref{tab:additional_pubmed}, we report the accuracy score of each model on PubMedQA before and after LoRA fine-tuning. Except LLM-Pruner, we see score improvement on all other models after fine-tuning. Thus, Table~\ref{tab:additional_pubmed} indicates that our fine-tuning is still improving model performance on PubMedQA in some ways.

\section{Experiments on BLOOM}
\label{sec:exp_bloom}
We conduct a small set of experiments in healthcare domain for illustrative purpose. SparseGPT is chosen for comparison, since it is the strongest baseline. We run SparseGPT under two settings: (1) only open-domain calibration dataset is used for pruning, and (2) both open-domain and domain-specific calibration datasets are used, which is the same as the setting in Section~\ref{sec:results}. All BLOOM experiments are based on the \texttt{bigscience/bloom-7b1} model on Hugging Face.

As shown in Table~\ref{tab:additional_bloom}, SparseGPT yields the best performance on BLOOM across all three metrics. Although \model{} surpasses SparseGPT on MedNLI when SparseGPT only uses open-domain data, it struggles on both medical perplexity and PubMedQA. Because our method is based on \citet{lv2023full} for fine-tuning and this fine-tuning pipeline only discusses performance scores on LLaMA, \citet{lv2023full} might require a significant adaptation when we change our backbone models from LLaMA to BLOOM. It might also not work well on BLOOM-based models when we integrate the general importance as a regularization term. Therefore, we might need to switch the fine-tuning pipeline we use in order to obtain the optimal performance of \model{}.

\section{Hyperparameters}
\label{sec:hyper}
We stick to the default values of hyperparameters for our baseline models. For \model{}, in the healthcare domain, we set $\lambda$ (regularization strength) and learning rate to $0.1$ and $0.03$. In the legal domain, we set $\lambda$ and learning rate to $0.001$ and $0.03$.

\begin{table}[t!]
\resizebox{\columnwidth}{!}{%
\begin{tabular}{lcc}
\toprule
Model & Before LoRA & After LoRA \\ \midrule
Dense & 39.20 & 64.60\\ 
Magnitude & 47.00 & 55.20 \\ 
LLM-Pruner & 51.40 & 40.20 \\ 
SparseGPT & 53.80 & 57.00 \\ 
\model{} & 58.80 & 59.20 \\ 

\bottomrule
\end{tabular}%
}
\caption{Accuracy scores of different models on PubMedQA dataset.}
\label{tab:additional_pubmed}
\end{table}

\begin{table}[t!]
\resizebox{\columnwidth}{!}{%
\begin{tabular}{lccc}
\toprule
Model & Perplexity & MedNLI & PubMedQA \\ \midrule
Dense & 9.40 & 33.26 & 23.72\\
SparseGPT* & 11.16 & 32.07 & 29.74 \\ 
SparseGPT & 10.88 & 33.47 & 24.23 \\ 
\model{} & 14.70 & 32.70 & 20.95\\ 

\bottomrule
\end{tabular}%
}
\caption{Performance of SparseGPT and \model{} (at 50\% sparsity)  on metrics of healthcare domain based on BLOOM. * denotes the model that only uses open-domain calibration data (C4) for pruning.}
\label{tab:additional_bloom}
\end{table}

\end{document}